\documentclass[10pt, a4paper]{article}
\usepackage{lrec2016}
\usepackage{multibib}
\newcites{languageresource}{Language Resources}
\usepackage{graphicx}

\usepackage{epstopdf}
\usepackage[latin1]{inputenc}
\usepackage{multirow}
\usepackage{color,soul} 
\usepackage{url} 
\usepackage{hyperref}

\title{Automatic Construction of Discourse Corpora for Dialogue Translation}

\name{Longyue Wang$^*$, Xiaojun Zhang$^*$, Zhaopeng Tu$^\dagger$, Andy Way$^*$, Qun Liu$^*$}

\address{$^*$ ADAPT Centre, School of Computing, Dublin City University, Ireland \\ $^\dagger$ Noah Ark Lab, Huawei Technologies, China \\
               \{lwang, xzhang, away, qliu\}@computing.dcu.ie, tu.zhaopeng@huawei.com }

\abstract{
In this paper, a novel approach is proposed to automatically construct parallel discourse corpus for dialogue machine translation. Firstly, the parallel subtitle data and its corresponding monolingual movie script data are crawled and collected from Internet. Then tags such as speaker and discourse boundary from the script data are projected to its subtitle data via an information retrieval approach in order to map monolingual discourse to bilingual texts. We not only evaluate the mapping results, but also integrate speaker information into the translation. Experiments show our proposed method can achieve 81.79\% and 98.64\% accuracy on speaker and dialogue boundary annotation, and speaker-based language model adaptation can obtain around 0.5 BLEU points improvement in translation qualities. Finally, we publicly release around 100K parallel discourse data with manual speaker and dialogue boundary annotation. \\ 
\newline 
 }

\begin{document}

\maketitleabstract

\section{Introduction}
Dialogue is an essential component of social behaviour to express human emotions, moods, attitudes and personality. To date, few researchers have investigated how to improve the machine translation (MT) of conversational material by exploiting their internal structure. This lack of research on the dialogue MT is a surprising fact, since dialogue exhibits more cohesiveness than single sentence and at least as much than textual discourse. 

Although there are a number of papers on corpus construction for various natural language processing (NLP) tasks, dialogue corpora are still scarce for MT. Some work regarding bilingual subtitles as parallel corpora exists, but it lacks rich information between utterances (sentence-level corpus) \cite{lavecchia2007building,tiedemann2007building,tiedemann2007improved,itamar2008using,tiedemann2008synchronizing,xiao2009constructing,tiedemann2012parallel,zhang2014dual}. Other work focuses on mining the internal structure in dialogue data from movie scripts. However, these are monolingual data which cannot used for MT \cite{danescu2011chameleons,banchs2012movie,walker2012annotated,schmitt2012parameterized}. In general, the fact is that bilingual subtitles are ideal resources to extract parallel sentence-level utterances, and movie scripts contain rich information such as dialogue boundaries and speaker tags.

Inspired by these facts, our initial idea was to build dialogue discourse corpus by bridging the information in these two kinds of resources (i.e., scripts and subtitles). The corpus should be parallel, align at the segment-level as well as contain rich dialogue information. We propose a simple but effective approach to build our dialogue corpus. Firstly, we extract parallel sentences from bilingual subtitles, and mine dialogue information from monolingual movie scripts. Secondly, we project dialogue information from script utterances to its corresponding parallel subtitle sentences using an information retrieval (IR) approach. Finally, we apply this approach to build a Chinese--English dialogue corpus, and also manually annotate dialogue boundaries and speaker tags based on automatic results.

To validate the effect of the proposed approach, we carried out experiments on the generated corpus. Experimental results show that the automatic annotation approach can achieve around 82\% and 98\% on speaker and dialogue boundaries annotation, respectively. Furthermore, we explore the integration of speaker information into MT via domain-adaptation techniques. Results show that we can improve translation performance by around 0.5 BLEU points compared to baseline system.

Generally, the contributions of this paper include the following:
\begin{itemize}
\item We propose an automatic method to build a segment-level dialogue parallel corpus with useful information, for building large-scale dialogue MT systems;
\item Through exploring dialogue information with MT, we show that speaker information is really helpful to dialogue MT systems;
\item We also manually annotate about 100K sentences from our dialogue corpus. The gold standard dataset\footnote{We release our DCU English-Chinese Dialogue Corpus in \url{http://computing.dcu.ie/~lwang/resource.html}.} can be further used to search for the coherence and consistency clues in discourse structure to implement a dialogue MT system.
\end{itemize}

The rest of the paper is organized as follows. In Section 2, we describe related work. Section 3 describes in detail our approaches to build a dialogue corpus as well as the structure of the generated database. The experimental results for both corpus annotation and translation are reported in Section 4. Finally, Section 5 presents our conclusions and future work plans.

\section{Related Work}

In the specific case of dialogue MT system, data acquisition can impose challenges including data scarcity, translation quality and scalability. The release of the Penn Discourse Treebank (PDTB)\footnote{Available at \url{https://www.seas.upenn.edu/\~pdtb}.} \cite{prasad2008penn} helped bring about a new sense of maturity in discourse analysis, finally providing a high-quality large-scale resource for training discourse parsers for English. Based on PDTB, some have applied the insights to MT \cite{meyer2012using}. A resource like the PDTB is extremely valuable, and it would be desirable to have a similar resource in dialogue or conversation as well. 

There are two directions of work related to dialogue corpus construction. One is parallel corpora construction for dialogue or conversation MT \cite{lavecchia2007building,tiedemann2007building,tiedemann2007improved,tiedemann2008synchronizing,itamar2008using,xiao2009constructing,tiedemann2012parallel}. Thanks to the effects of crowdsourcing and fan translation in audiovisual translation \cite{ol2012fan}, we can regard subtitles as parallel corpora. \newcite{zhang2014dual} leveraged the existence of bilingual subtitles as a source of parallel data for the Chinese-English language pair to improve the MT systems in the movie domain. However, their work only considers sentence-level data instead of extracting more useful information for dialogues. Besides, Japanese researchers constructed a speech dialogue corpus for a machine interpretation system \cite{aizawa2000spoken,matsubara2002bilingual,ryu2003bilingual,takezawa2003collecting}. They collected speech dialogue corpora for machine interpretation research via recording and transcribing Japanese/English interpreters' consecutive/simultaneous interpreting in the booth. The German VERBMOBIL speech-to-speech translation programme \cite{wahlster2013verbmobil} also collected and transcribed task-oriented dialogue data. This related work focused on speech-to-speech translation including three modules of automatic speech recognition (ASR), MT and text-to-speech(TTS).

The other one is mining rich information from other resources such as movie scripts. \newcite{danescu2011chameleons} created a conversation corpus containing large metadata-rich collections of fictional conversations extracted from raw movie scripts. Both \newcite{banchs2012movie} and CMU released dialogue corpora extracted from the Internet Movie Script Database (IMSDb).\footnote{Available at \url{http://www.imsdb.com}.} 
Based on IMSDb, \newcite{walker2012annotated} annotated 862 film scripts to learn and characterize the character style for an interactive story system, and \newcite{schmitt2012parameterized} annotated 347 dialogues to explore a spoken dialogue system. The resource of movie scripts, such as IMSDb, is good enough to generate conversational discourse for dialogue processing. However, monolingual movie scripts are not enough for MT which needs a large-scale bilingual dialogue corpus to train and tune translation models.

\section{Building A Parallel Dialogue Corpus}

As already stated, our presented parallel dialogue corpus is extracted from bilingual movie/episode subtitles and monolingual scripts. We extend previous work on movie scripts to scripts of TV series such as \textit{Friends}. From IMSDb and SimplyScripts\footnote{Available at \url{http://www.simplyscripts.com}.} and the like, we crawled movie/episode scripts data. In addition, we collected the English-Chinese bilingual subtitles from multiple audiovisual translation web resources such as Shooter\footnote{Available at \url{http://sub.makedie.me}.} and Opensubtitles.~\footnote{Available at \url{http://www.opensubtitles.org}.} Based on the hypothesis that  both a script and a subtitle exist for the same movie or episode, the method can be described in a pipeline as follows:
\begin{itemize}
\item [(1)] given a monolingual movie/episode script, we identify dialogue boundaries and speaker tags using clues such as format and story structure tags in the script;
\item [(2)] for a bilingual subtitle, we align each sentence with its translation using clues such as format and time information;
\item [(3)] for each utterance in a processed script, we apply IR techniques to match it with the line(s) in its corresponding processed subtitle according to the shared language;
\item [(4)] for each matched term, we map the useful annotations such as speaker and dialogue boundaries from the script side to the matched line(s) in its subtitle side. 
\end{itemize}

\subsection{Script and Subtitle}
\begin{figure*}
\graphicspath{ {figures/} }
\begin{center}
\includegraphics[scale=0.4]{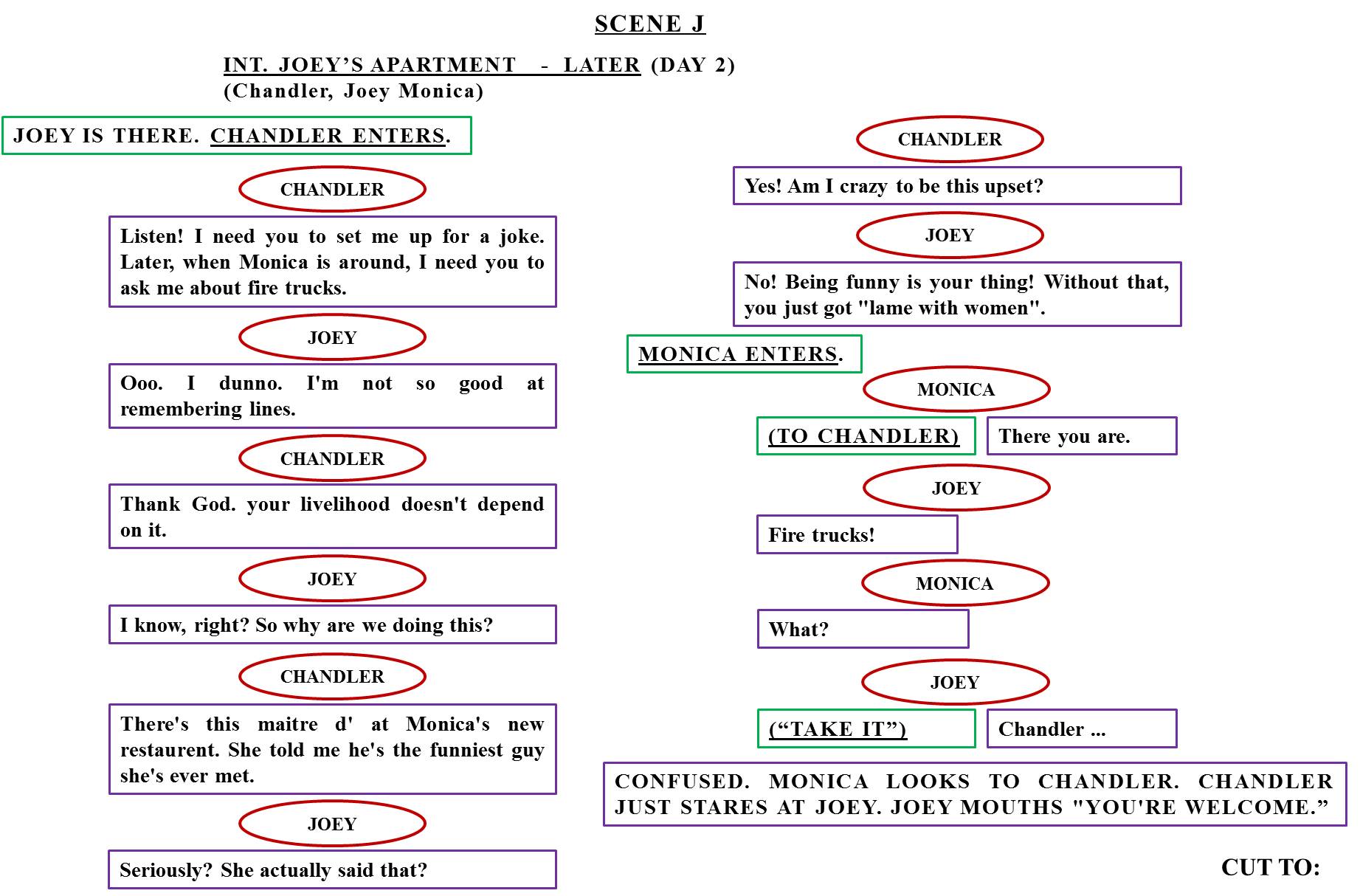}
\caption{Examples of scripts of \textit{Friends} in English}
\label{fig.1}
\end{center}
\end{figure*}

Figure~\ref{fig.1} depicts a browser snapshot illustrating an episode script layout of \textit{Friends}. There are three kinds of information: speaker, shot/scene and action information in the script. The speaker element (red ellipses) contains the corresponding character who says the utterance(s). The shot/scene tags (e.g., ``SCENE'', ``SHOT'', ``CUT INTO:'' and ``CUT TO:'' etc.) can be regarded as the boundaries of dialogues. For instance, the tags ``SCENE J'' and ``CUT TO:'' refer to the beginning and end of a dialogue, respectively. The action (green frames) contains all additional information of a narrative nature and explains what is happening in the scene. In this work, we focus on the first two information type while ignoring final one. 

\begin{figure*}
\graphicspath{ {figures/} }
\begin{center}
\includegraphics[scale=0.6]{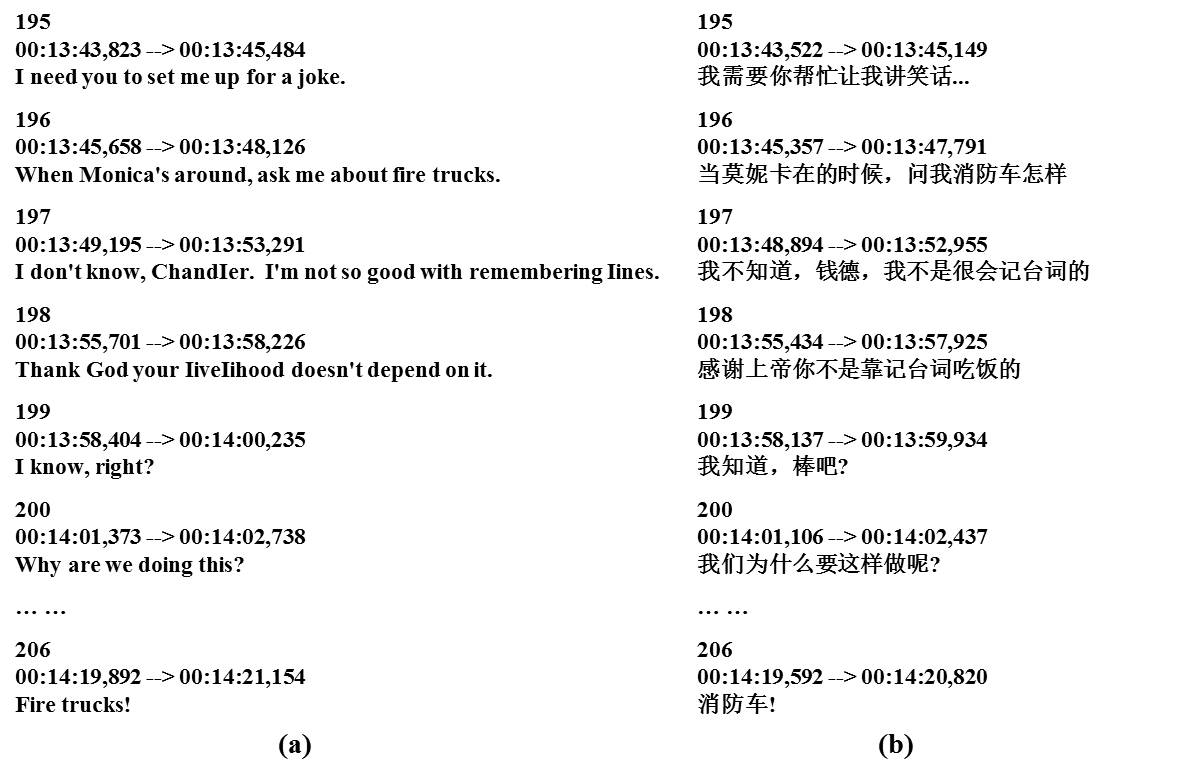}
\caption{Examples of bilingual subtitles (SRT) of \textit{Friends} in English and Chinese}
\label{fig.2}
\end{center}
\end{figure*}

Figure~\ref{fig.2} is the corresponding bilingual subtitle of the script in Figure~\ref{fig.1}. Subtitles are often organized in two formats: Advanced SubStation Alpha (ASS) and SubRip Text (SRT). As most lines are one-to-one aligned on two language sides, it easy to process them into a parallel corpus. We also use line id and time line information to deal with one-to-many or mismatching cases.

Based on the above rules, we extract useful information from both scripts and subtitles. In order to obtain high-quality data, we also apply a series of techniques including language detection, simplified-traditional Chinese conversation, coding conversation and punctuation normalization. After processing scripts and subtitles, the next step is to match and project terms from script side to subtitle side. 

\subsection{Matching and Projection}
Comparing examples in Figures~\ref{fig.1} and \ref{fig.2}, we found that the script and the subtitle share the same language (i.e., English). However, subtitle lines are not always the same as the utterances in a script for the actors may change their lines on site, either slightly or to a greater extent. For example, the first utterance in the script is \textit{Later, when Monica's around, I want you to ask me about fire trunks} while the corresponding line in the subtitle is \textit{When Monica's around, ask me about fire trunks.}. Another phenomenon is that one utterance on script side may be split into several lines on subtitle side. This change is made to accommodate the size of the TV screen. It is a big challenge to deal with these changed, missing or duplicated terms during matching. All the above problems make the task a complex $N$-to-$N$ matching where $N\geq0$.

Therefore, we regard the matching and projection as an IR task~\cite{wang2012improvement}. The Vector Space Model (VSM) \cite{salton1975vector} is a state-of-the-art IR model in which each document is represented as a vector of identifiers (here we describe each identifier as a term). The $i$th utterance $D_i$ in the script is represented as a vector $D_i = [w_{1,i}, w_{2,i}, ... w_{k,i}]$, in which $k$ is the size of the term vocabulary. Many similarity functions can be employed to calculate the similarity between two utterance vectors \cite{Cha07comprehensivesurvey}. Here we apply the cosine distance:
\begin{eqnarray}
\textit{sim}(d_i, d_j) = {\sum_{k=1}^{N} w_{i,k} \cdot w_{j,k} } { \sqrt {\sum_{k=1}^{N} w_{i,k} } \cdot {\sqrt {\sum_{k=1}^{N} w_{j,k} } } }
\end{eqnarray}
where $N$ is the number of terms in an utterance vector, and $w_{i,k}$ and $w_{j,k}$ represent the weight of the $i\textit{th}$/$j\textit{th}$ term in the utterance $D_{i}$/$D_{j}$ respectively. Technically, the distance between documents in VSM is calculated by comparing the deviation of angles between vectors. A Boolean Retrieval Model sets a term weight to be either 0 or 1, while an alternative solution is calculating the term weights according to the appearance of a term within the document collection.

To calculate the term weights according to the appearance of a term within the document collection, we apply term frequency-inverse document frequency (\textit{TF-IDF}) \cite{ramos2003using} as one term-weighting model. The weight $w$ of each term $t$ is determined by its own term frequency $tf(t, d)$ in a document $d$ and its inverse document frequency $idf(t, d, D)$ within the search collection. The definition of term weight $w_{t, d}$ is shown as in Eq. (2) and (3):
\begin{eqnarray}
w_{t, d} &=& tf(t,d) \cdot idf(t, d, D) \\
\label{eq:3:idf}
\textit{idf}(t, d, D) &=& \log \big( \frac{|D|}{| \{ d \in D | t \in d \} |} \big)
\end{eqnarray}
where $D$ is the total number of documents in the document collection. 

\begin{figure*}
\graphicspath{ {figures/} }
\begin{center}
\includegraphics[scale=0.7]{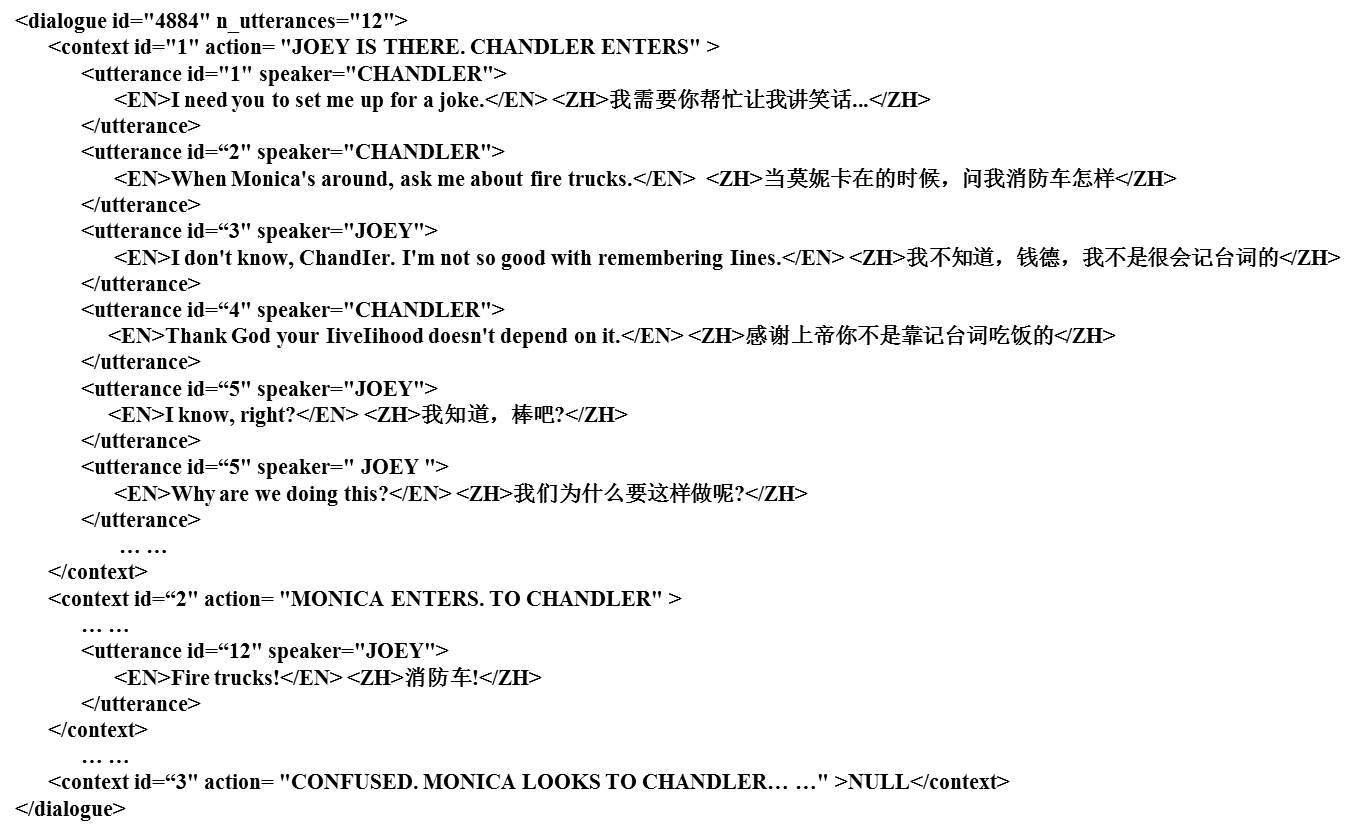}
\caption{A sample of generated XML of dialogue in episode script}
\label{fig.3}
\end{center}
\end{figure*}

In practice, we regard each utterance as a document and build the index for each movie script. Then we use each subtitle sentence as a query to search for target related utterances. In order to deal with inconsistency problems, we employ several strategies:
\begin{itemize}
\item For better indexing and searching, we split the sentences/utterances into the smallest units using a sentence splitter;
\item Except for punctuation mark, we do not remove any stop words. Furthermore we low-case each word;
\item For each original query, it can be split into $n$ sub-queries. For each sub-query, we apply $1$-best search. Then search results of the sub-queries are combined to vote for the best candidate for the original query.
\item One query may be similar to several utterances in different lines of a script. The candidate closest to the last matched term is more likely to be the right answer. Thus we impose a dynamic window for subspace searching.
\end{itemize}
After the script and subtitle are bridged, we project speaker tags and dialogue boundaries in scripts to their corresponding lines in subtitles. Finally, we preserve the results in XML format, which is illustrated in Figure~\ref{fig.3}.

\section{Experiments and Results}
For dialogue corpus construction, we apply our methods to a ten-season sitcom \textit{Friends}. We extract and process both scripts and subtitles of \textit{Friends} (described in Section 3.1) and then bridge them (described in Section 3.2) to build a dialogue corpus in the format of Figure~\ref{fig.3}. For data processing, we employ the sentence splitter and English tokenizer in the Moses toolkit and our in-house Chinese segmentor \cite{wang2012crfs}. Furthermore, we employ Apache Lucene\footnote{Available at \url{https://lucene.apache.org}.} for indexing and search tasks. Table~\ref{tab.1} presents the main statistics of the resulting bilingual dialogue corpus. We obtained 5,428 bilingual dialogues with annotated speaker and dialogue boundary information.
\begin{table}[!h]
\begin{center}
\begin{tabular}{|l|l|}
      \hline
      Item&Size\\
      \hline\hline
      Total number of scripts processed & 236\\
      Total number of dialogues & 5,428\\
      Total number of speakers & 42\\
      Total number of utterances & 109,268\\
      Average amount of dialogues per script & 23 \\
      Average amount of speakers per dialogue & 3.5 \\
      Average amount of utterances per dialogue & 20 \\
      \hline
\end{tabular}
\caption{Statistics of generated parallel dialogue corpus}
\label{tab.1}
\end{center}
\end{table}

To verify the validity of our methods (described in Section 3), we conduct an evaluation on the matching accuracy of speaker tags and dialogue boundaries in the generated corpus. To generate gold standard reference, we also manually annotate the dialogue information based on the generated parallel dialogue corpus. The agreements between automatic labels and manual labels is 81.79\% on speaker and 98.64\% on dialogue boundary, respectively. This indicates that the proposed automatic annotation strategy through mapping is reasonably trustworthy.

Furthermore, we conduct a simple experiment to explore the effects of speaker tags on dialogue MT. We first build a baseline MT engine using Moses~\cite{Koehn:ACL:2007} on our generated parallel corpus (described in Table~\ref{tab.1}). We train a $5$-gram language model (LM) using the SRI Language Toolkit~\cite{Stolcke:2002:CSLP} on the target side of parallel corpus. Besides, we use GIZA++~\cite{Och:2003} for word alignment and minimum error rate training~\cite{Och:2003b} to optimize feature weights. Based on the hypothesis that different types of speakers may have specific speaking styles, we employ a language model adaptation method to boost the MT system \cite{wang2014combining}. Instead of building a LM on the whole data, we split the data into two separate parts according the speakers' sex and then build two separate LMs. As Moses supports multiple LM integration, we directly feed Moses two LMs. The translation results are listed in Table~\ref{tab.2}.
\begin{table}[!h]
\begin{center}
\begin{tabular}{|l|l|l|l|}
     \hline
     Systems & Lang. & Dev Set & Test Set\\
    \hline\hline
        \multirow{2}{*}{ZH-EN} &    Baseline  & 20.32 & 16.33\\
        &   Speaker$_{LM}$  & 21.05 & 16.83 (+0.50)\\
    \hline\hline
        \multirow{2}{*}{EN-ZH} & Baseline   & 16.78 & 14.11\\
        &   Speaker$_{LM}$  & 17.23 & 14.54 (+0.43)\\
     \hline
\end{tabular}
\caption{Translation results on speaker based language model adaption.}
\label{tab.2}
\end{center}
\end{table}
For Chinese-to-English (i.e. ``ZH-EN''), the baseline system achieves 20.32 and 16.33 in BLEU score on development and test data, respectively, while for English-to-Chinese (i.e. ``EN-ZH''), the scores are 16.78 and 14.11 in BLEU score. The BLEU scores are relatively low because 1) we have only one reference, 2) the training corpus is small, and 3) dialogue MT is a challenging task. By using LM adaptation, we improve the performance on test data by +0.50 and +0.43 BLEU points on Chinese-to-English and English-to-Chinese tasks respectively.

\section{Conclusions and Future Work}
We propose a novel approach to build a parallel dialogue discourse corpus from monolingual scripts and their corresponding bilingual subtitles. We identify the dialogue boundaries according to the scene or shot tags in the script to segment the monolingual dialogue, and then map the matched monolingual dialogues to the source part of the bilingual subtitles with the speaker and utterance elements in order to obtain the bilingual discourse dialogues. Finally we align the bilingual dialogue subtitle lines to produce suitable MT training material. 

We expand the current dialogue generation resources from movie scripts to movie/episode scripts, and specify the current parallel corpus construction to bilingual dialogue corpus building based on bilingual subtitles. We pilot this approach on a 10-season sitcom \textit{Friends} and automatically generated 5,428 bilingual parallel dialogue discourses. This is a quick way to generate a bilingual dialogue corpus. 

To validate the effect of the proposed approach, we annotated the speaker and dialogue boundary elements manually in 4-season \textit{Friends} data and compared the manual results with our automatic findings. Experimental results show that the automatic annotation approach can achieve around 81.79\% and 98.64\% on dialogue boundaries and speaker tags, respectively. Furthermore, we explore the integration of speaker tags into MT using domain-adaptation techniques. The experiments show that we can improve translation performance compared to a baseline system.  

As far as future work is considered, we intend to explore automatic dialogue detection from bilingual subtitles. A reachable goal is to utilize the resulting bilingual dialogue corpus based on our approach to also summarize the discourse elements such as coherence and co-reference, speaker relationship and time information of the subtitle lines. Some supervised and semi-supervised methods and machine learning approaches can be used on these tasks.

\section{Acknowledgements}

This work is supported by the Science Foundation of Ireland (SFI) ADAPT project (Grant No.:13/RC/2106), and partly supported by the DCU-Huawei Joint Project (Grant No.:201504032-A (DCU), YB2015090061 (Huawei)). It is partly supported by the Open Projects Program of National Laboratory of Pattern Recognition (Grant 201407353) and the Open Projects Program of Centre of Translation of GDUFS (Grant CTS201501).

\section{Bibliographical References}
\bibliographystyle{lrec2016}
\bibliography{xample}

\end{document}